\documentclass[10pt,twocolumn,letterpaper]{article}

\usepackage{cvpr}
\usepackage{times}
\usepackage{epsfig}
\usepackage{graphicx}
\usepackage{amsmath}
\usepackage{amssymb}

\usepackage{booktabs}
\usepackage{pifont} %
\usepackage{comment}

\usepackage{pgfplots} %
\usepackage{pgfplotstable} %
\pgfplotsset{compat=newest} %
\usetikzlibrary{plotmarks} %
\usetikzlibrary{colorbrewer} %

\usepackage[pagebackref=true,breaklinks=true,letterpaper=true,colorlinks,bookmarks=false]{hyperref}

\cvprfinalcopy %

\def\cvprPaperID{3335} %

\newcommand{\PAR}[1]{\vskip1pt \noindent {\bf #1~}}
\newcommand{\PARbegin}[1]{\noindent {\bf #1~}}

\ifcvprfinal\fi
\begin{document}

\title{FEELVOS: Fast End-to-End Embedding Learning for Video Object Segmentation}

\author{
  \hspace{-1.3cm}
  \begin{tabular}[t]{c}
    Paul Voigtlaender$^{1,*}$, Yuning Chai$^{2,\dagger}$, Florian Schroff$^2$, Hartwig Adam$^2$,\\  Bastian Leibe$^1$, Liang-Chieh Chen$^2$ \\
    RWTH Aachen University$^1$, Google Inc.$^2$\\
    {\tt\small voigtlaender@vision.rwth-aachen.de}
\end{tabular}
}

\maketitle

\ifcvprfinal
\let\thefootnote\relax\footnotetext{$^*$Work done during an internship at Google Inc.}\footnotetext{$\dagger$ Now at Waymo LLC.}
\thispagestyle{empty}
\fi

\begin{abstract}
   Many of the recent successful methods for video object segmentation (VOS) are overly complicated, heavily rely on fine-tuning on the first frame, and/or are slow, and are hence of limited practical use. In this work, we propose FEELVOS as a simple and fast method which does not rely on fine-tuning. In order to segment a video, for each frame FEELVOS uses a semantic pixel-wise embedding together with a global and a local matching mechanism to transfer information from the first frame and from the previous frame of the video to the current frame. In contrast to previous work, our embedding is only used as an internal guidance of a convolutional network. Our novel dynamic segmentation head allows us to train the network, including the embedding, end-to-end for the multiple object segmentation task with a cross entropy loss. We achieve a new state of the art in video object segmentation without fine-tuning with a $\mathcal{J}$\&$\mathcal{F}$ measure of $71.5\%$ on the DAVIS 2017 validation set. We make our code and models available at \url{https://github.com/tensorflow/models/tree/master/research/feelvos}.
\end{abstract}

\section{Introduction}
\label{sec:intro}

Video object segmentation (VOS) is a fundamental task in computer vision, with important applications including video editing, robotics, and self-driving cars. In this work, we focus on the semi-supervised VOS setup in which the ground truth segmentation masks of one or multiple objects are given for the first frame in a video. The task is then to automatically estimate the segmentation masks of the given objects for the rest of the video. With the recent advances in deep learning and the introduction of the DAVIS datasets \cite{DAVIS2016,DAVIS2017}, there has been tremendous progress in tackling the semi-supervised VOS task. However, many of the most successful methods rely on fine-tuning of the model using the first-frame annotations and have very high runtimes which are not suitable for most practical applications. Additionally, several successful methods rely on extensive engineering, resulting in a high system complexity with many components. For example, the 2018 DAVIS challenge was won by PReMVOS \cite{Luiten18ACCV, Luiten18DAVIS, Luiten18ECCVW} which employs four different neural networks together with fine-tuning and a merging algorithm resulting in a total runtime of around 38 seconds per video frame. While it delivered impressive results, the practical usability of such algorithms is limited.

\begin{table}
\begin{centering}
\scalebox{0.96}{
\begin{tabular}{ccccc}
\toprule 
 & Simple & Fast & End-to-end & Strong\tabularnewline
\midrule
PML \cite{Chen18CVPR} & \ding{51} & \ding{51} &  & \tabularnewline
OSMN \cite{Yang18CVPR} & \ding{51} & \ding{51} &  & \tabularnewline
FAVOS \cite{Cheng18CVPR} & \ding{51} & \ding{51} &  & \tabularnewline
VideoMatch \cite{Hu18ECCV} & \ding{51} & \ding{51} & \ding{51} & \tabularnewline
RGMP \cite{Oh18CVPR} &  & \ding{51} &  & \ding{51}\tabularnewline
FEELVOS (ours) & \ding{51} & \ding{51} & \ding{51} & \ding{51}\tabularnewline
\midrule 
PReMVOS \cite{Luiten18ACCV} &  &  &  & \ding{51}\tabularnewline
OnAVOS \cite{voigtlaender17BMVC} & \ding{51} &  &  & \tabularnewline
\bottomrule
\end{tabular}
}
\par\end{centering}
\caption{\label{tab:design-goals}Design goals overview. The table shows which
of our design goals (described in more detail in the text) are achieved
by recent methods. Our method is the only one which %
 fulfills all design goals.}
\end{table}

\begin{figure*}[t]
\centering
\includegraphics[width=\textwidth]{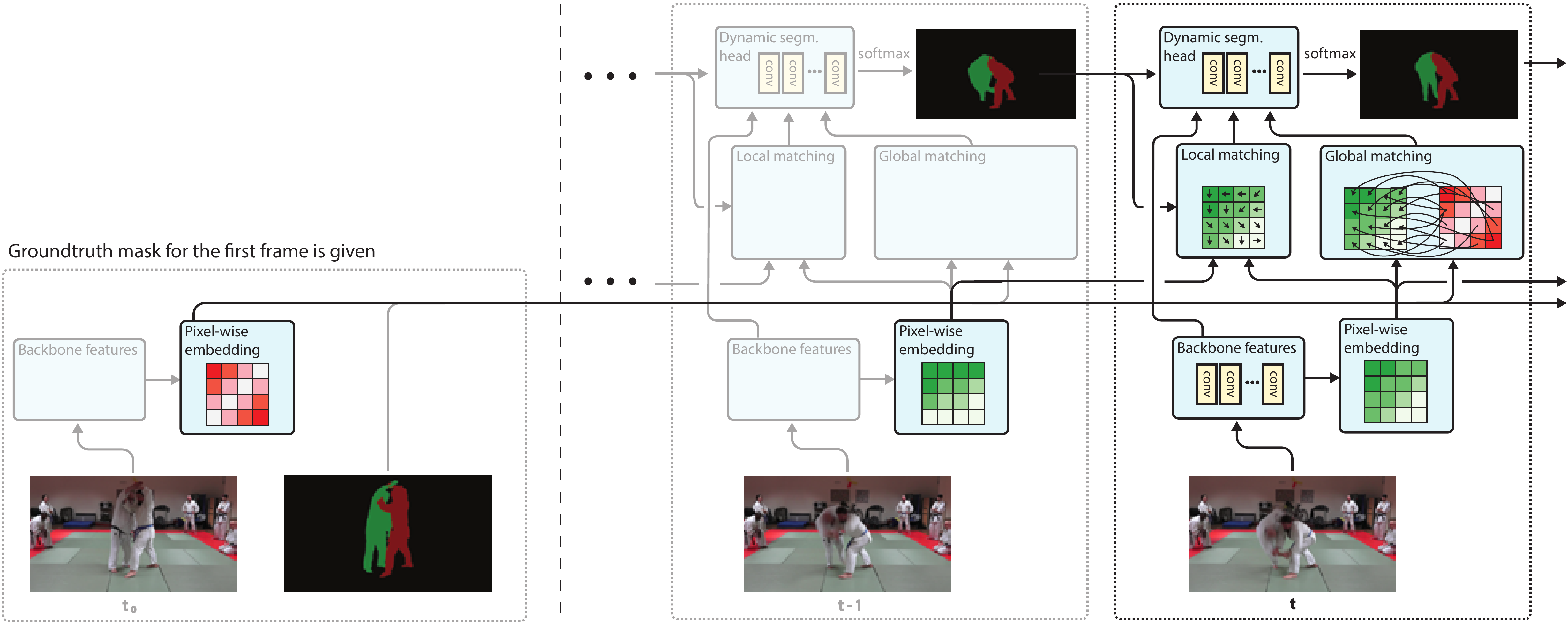}
\caption{\label{fig:method_overview}Overview of the proposed FEELVOS method. In order to segment the image of the current frame, backbone features and pixel-wise embedding vectors are extracted for it. Afterwards the embedding vectors are globally matched to the first frame and locally matched to the previous frame to produce a global and a local distance map. These distance maps are combined with the backbone features and the predictions of the previous frame and then fed to a dynamic segmentation head which produces the final segmentation. For details about handling of multiple objects see Fig.~\ref{fig:seg_head}.
}
\end{figure*}

\PAR{Design Goals.} In order to ensure maximum practical usability, in this work, we develop a method for video object segmentation with the following design goals: A VOS method should be
\begin{itemize}
\vspace{-5pt}
\setlength\itemsep{-0.4em}
\item{Simple: Only a single neural network and no simulated data is used.}
\item{Fast: The whole system is fast for deployment. In particular, the model does not rely on first-frame fine-tuning.}
\item{End-to-end: The multi-object segmentation problem, where each video contains a different number of objects, is tackled in an end-to-end way.}
\item{Strong: The system should deliver strong results, with more than $65\%$ $\mathcal{J}$\&$\mathcal{F}$ score on the DAVIS 2017 validation set.}
\end{itemize}
Table~\ref{tab:design-goals} shows an overview of which of our design goals are achieved by current methods that do not use fine-tuning, and by PReMVOS \cite{Luiten18ACCV, Luiten18DAVIS, Luiten18ECCVW} and OnAVOS \cite{voigtlaender17BMVC} as examples for methods with fine-tuning. For a more detailed discussion of the individual methods, see Section~\ref{sec:related-work}.

Towards building a method which fulfills all design goals, we take inspiration from the %
Pixel-Wise Metric Learning (PML) \cite{Chen18CVPR} method. PML learns a pixel-wise embedding using a triplet loss and at test time assigns a label to each pixel by nearest neighbor matching in pixel space to the first frame. PML fulfills the design goals of being simple and fast, but does not learn the segmentation in an end-to-end way and often produces noisy segmentations due to the hard assignments via nearest neighbor matching.

We propose Fast End-to-End Embedding Learning for Video Object Segmentation (FEELVOS) to meet all of our design goals (see Fig.~\ref{fig:method_overview} for an overview). Like PML \cite{Chen18CVPR}, FEELVOS uses a learned embedding and nearest neighbor matching, but we use this mechanism as an internal guidance of the convolutional network instead of using it for the final segmentation decision. This allows us to learn the embedding in an end-to-end way using a standard cross entropy loss on the segmentation output. By using the nearest neighbor matching only as a soft cue, the network can recover from partially incorrect nearest neighbor assignments and still produce accurate segmentations. We achieve a new state of the art for multi-object segmentation without fine-tuning on the DAVIS 2017 validation dataset with a $\mathcal{J}$\&$\mathcal{F}$ mean score of $71.5\%$. %

\section{Related Work}\label{sec:related-work}

\PARbegin{Video Object Segmentation with First-Frame Fine-tuning.}
Many approaches for semi-supervised video object segmentation rely on fine-tuning using the first-frame ground truth. 
OSVOS \cite{OSVOS} uses a convolutional network, pre-trained for foreground-background segmentation, %
and fine-tunes it on the first-frame ground truth of the target video at test time. 
OnAVOS \cite{voigtlaender17BMVC, voigtlaender17DAVIS} and OSVOS-S \cite{Maninis18TPAMI} extend OSVOS by an online adaptation mechanism, and by semantic information from an instance segmentation network, respectively.
Another approach is to learn to propagate the segmentation mask from one frame to the next using optical flow as done by MaskTrack \cite{masktrack}.  %
This approach is extended by LucidTracker \cite{lucidtracker} which introduces an elaborate data augmentation mechanism. 
Hu \etal \cite{Hu18CVPR} propose a motion-guided cascaded refinement network which  works on a coarse segmentation from an active contour model. %
MaskRNN \cite{Hu17NIPS} uses a recurrent neural network to fuse the output of two deep networks. 
Location-sensitive embeddings used to refine an initial foreground prediction are explored in LSE \cite{Ci18ECCV}.
MoNet \cite{Xiao18CVPR} exploits optical flow motion cues by feature alignment and a distance transform layer.
Using reinforcement learning to estimate a region of interest to be segmented is explored by 
Han \etal \cite{Han18CVPR}. %
DyeNet \cite{Li18ECCV} uses a deep recurrent network which combines a temporal propagation and a re-identification module. 
PReMVOS \cite{Luiten18ACCV, Luiten18DAVIS, Luiten18ECCVW} combines four different neural networks together with extensive fine-tuning and a merging algorithm and won the 2018 DAVIS Challenge \cite{Caelles18Arxiv} and also the 2018 YouTube-VOS challenge \cite{Xu18ECCV}.

Despite achieving impressive results, all previously mentioned methods do not meet the design goal of being fast, since they rely on fine-tuning on the first frame.

\PAR{Video Object Segmentation without First-Frame Fine-tuning.}
While there is a strong focus in semi-supervised VOS on exploiting the first-frame information by fine-tuning, there are some recent works which  aim to achieve a better runtime and usability by avoiding fine-tuning. 
OSMN \cite{Yang18CVPR} combines a segmentation network with a modulator, which manipulates intermediate layers of the segmentation network without requiring fine-tuning. 
FAVOS \cite{Cheng18CVPR} uses a part-based tracking method to obtain bounding boxes for object parts and then produces segmentation masks using a region-of-interest based segmentation network.
The main inspiration of the proposed FEELVOS approach is PML \cite{Chen18CVPR}, which uses a pixel-wise embedding learned with a triplet loss together with a nearest neighbor classifier.
VideoMatch \cite{Hu18ECCV} uses a soft matching layer which is very similar to PML and considers for each pixel in the current frame the closest $k$ nearest neighbors to each pixel in the first frame in a learned embedding space. Unlike PML, it directly optimizes the resulting segmentation instead of using a triplet loss. %
However, the final segmentation result is still directly derived from the matches in the embedding space which makes it hard to recover from incorrect matches. In our work, we use the embedding space matches only as a soft cue which is refined by further convolutions.

OSMN \cite{Yang18CVPR}, FAVOS \cite{Cheng18CVPR}, PML \cite{Chen18CVPR}, and VideoMatch \cite{Hu18ECCV} all achieve very high speed and effectively bypass fine-tuning, but we show that the proposed FEELVOS produces significantly better results.

RGMP \cite{Oh18CVPR} uses a Siamese encoder with two shared streams. The first stream encodes the video frame to be segmented together with the estimated segmentation mask of the previous frame. The second stream encodes the first frame of the video together with its given ground truth segmentation mask. The features of both streams are then concatenated and combined by a global convolution block and multiple refinement modules to produce the final segmentation mask. The architecture of RGMP has similarities with ours, in particular both RGMP and FEELVOS use both the first and the previous video frame images and segmentation masks as information which is exploited inside the network. However, RGMP combines these sources of information just by stacking together features, while we employ a feature-based matching mechanism inspired by PML \cite{Chen18CVPR} which allows us to systematically handle multiple objects in an end-to-end way.
Like FEELVOS, RGMP does not require any fine-tuning, is fast, and achieves impressive results. However, RGMP does not meet the design goal of tackling the multi-object segmentation task in an end-to-end way. %
The whole network needs to be run once for each %
object and the multi-object segmentation is performed by heuristic merging. %
Additionally RGMP does not fulfill the design goal of being simple, since it relies on an elaborate training procedure which involves multiple datasets, synthetic data generation and backpropagation through time. We show that despite using a much simpler training procedure and no simulated data, FEELVOS produces better results than RGMP.

\PAR{Instance Embedding Learning.}
Li \etal \cite{Li18CVPR} propose an embedding based method for performing video object segmentation in an unsupervised way, \ie, without using any ground truth information from the first frame.
For image instance segmentation, Fathi \etal \cite{Fathi17Arxiv} propose learning an instance embedding for each pixel. We adopt the same embedding distance formulation, but do not use their proposed seed points and we train the embedding in a different way. %

\section{Method}

\PARbegin{Overview.} We propose FEELVOS for fast semi-supervised video object segmentation. FEELVOS uses a single convolutional network and requires only a single forward pass for each video frame. See Fig.~\ref{fig:method_overview} for an overview of FEELVOS.
The proposed architecture uses DeepLabv3+ \cite{Chen18ECCV} (with its output layer removed) as its backbone to extract features with a stride of $4$. On top of that, we add an embedding layer which extracts embedding feature vectors at the same stride.
Afterwards, for each object we compute a distance map by globally matching the embedding vectors of the current frame to the embedding vectors belonging to this object in the first frame.
Additionally, we use the predictions of the previous time frame in order to compute for each object another distance map by locally matching the current frame embeddings to the embedding vectors of the previous frame. Both global and local matching will be described in more detail below.
Like in MaskTrack \cite{masktrack} or RGMP \cite{Oh18CVPR} we also use the predictions of the previous frame directly as an additional cue.
Finally, we combine all available cues, \ie, the global matching distance maps, the local matching distance maps, the predictions from the previous frame, and the backbone features. We then feed them to a dynamic segmentation head which produces for each pixel (with a stride of $4$) a posterior distribution over all objects which are present in the first frame. The whole system is trained end-to-end for multi-object segmentation without requiring a direct loss on the embedding. In the following, we will describe each of the components in more detail.

\PAR{Semantic Embedding.}
For each pixel $p$, we extract a semantic embedding vector $e_p$ in the learned embedding space. The idea of the embedding space is that pixels belonging to the same object instance (in the same frame or in different frames) will be close in the embedding space and pixels which belong to distinct objects will be far away. Note that this is not explicitly enforced, since instead of using distances in the embedding space directly to produce a segmentation like in PML \cite{Chen18CVPR} or VideoMatch \cite{Hu18ECCV}, we use them as a soft cue which can be refined by the dynamic segmentation head.
However, in practice the embedding indeed behaves in this way since this delivers a strong cue to the dynamic segmentation head for the final segmentation.

Similar to Fathi \etal \cite{Fathi17Arxiv}, we define the distance between pixels $p$ and $q$ in terms of their corresponding embedding vectors $e_p$ and $e_q$ by
\begin{equation}
  d(p,q) = 1-\frac{2}{1+\exp(\lVert e_{p}-e_{q}\rVert^{2})}.
\end{equation}
The distance values are always between $0$ and $1$. For identical pixels, the embedding distance is $d(p,p)=1-\frac{2}{1+\exp(0)}=0$, and for pixels which are very far away in the embedding space, we have $d(p,q)=1-\frac{2}{1+\exp(\infty)}=1.$

\PAR{Global Matching.}
\begin{figure}
\centering
\includegraphics[width=0.47\textwidth]{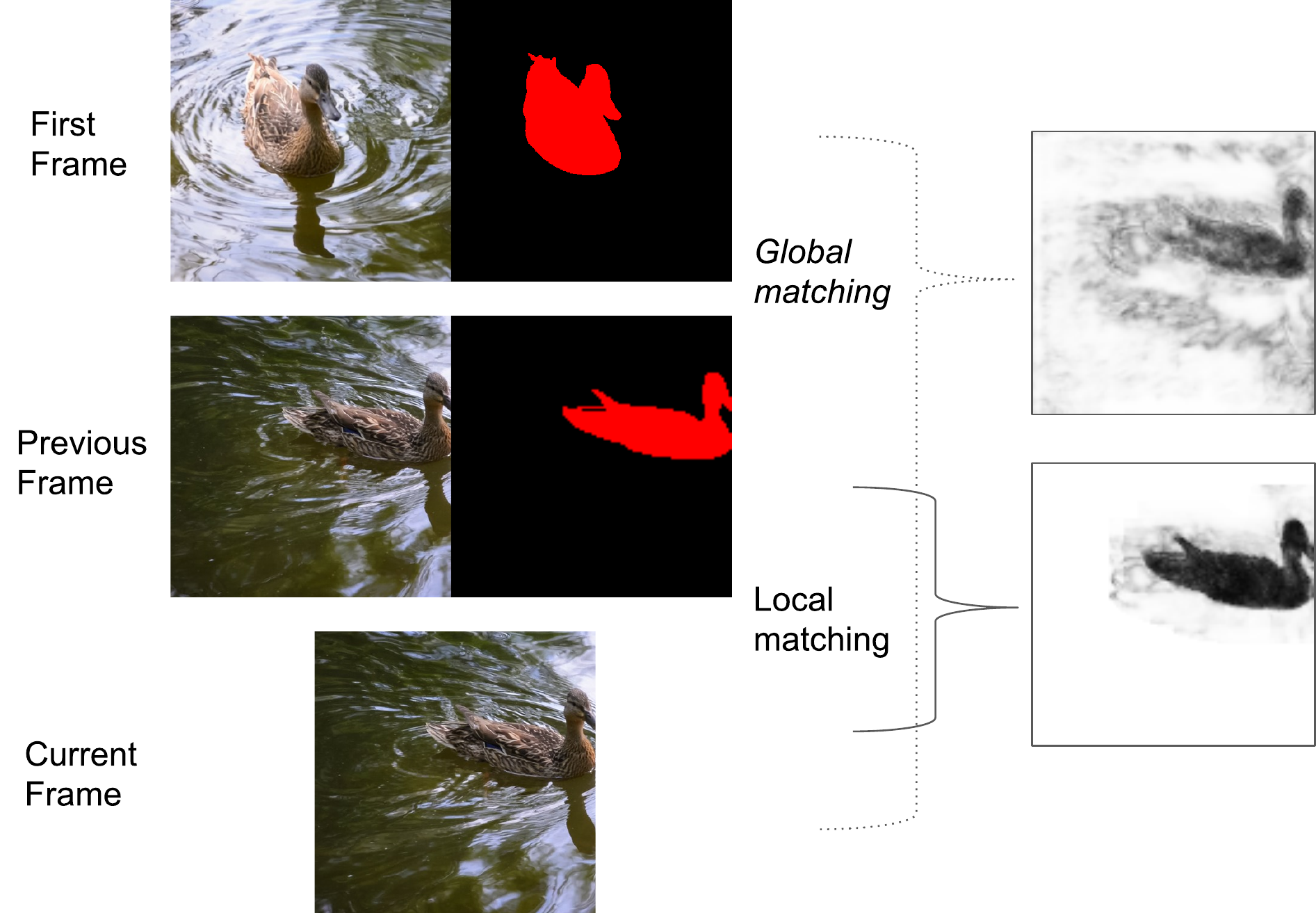}
\caption{\label{fig:matching} Global and local matching. For a given object (in this case the duck), global matching matches the embedding vectors of the current frame to the embedding vectors of the first frame which belong to the object and produces a distance map. Dark color denotes low distances. Note that the global distance map is noisy and contains false-positives in the water.
Local matching is used to match the current frame embeddings to the embeddings of the previous frame which belong to the object. For local matching, matches for a pixel are only allowed in a local window around it.%
}
\end{figure}
Similar to PML \cite{Chen18CVPR} and VideoMatch \cite{Hu18ECCV}, we transfer semantic information from the first video frame for which we have the ground truth to the current frame to be segmented by considering nearest neighbors in the learned embedding space.

Let $\mathcal{P}_t$ denote the set of all pixels (with a stride of $4$) at time $t$ and $\mathcal{P}_{t,o} \subseteq \mathcal{P}_{t}$ the set of pixels at time $t$ which belong to object $o$. 
We then compute the global matching distance map $G_{t,o}(p)$ for each ground truth object $o$ and each pixel $p\in \mathcal{P}_t$ of the current video frame $t$ as the distance to its nearest neighbor in the set of pixels $\mathcal{P}_{1,o}$ of the first frame which belong to object $o$, \ie,
\begin{equation}
  G_{t,o}(p) = \min_{q \in \mathcal{P}_{1,o}} d(p,q).
\end{equation}
Note that $\mathcal{P}_{1,o}$ is never empty, since we consider exactly the objects $o$ which are present in the first frame, and note that background is handled like any other object. $G_{t,o}(p)$ provides for each pixel and each object of the current frame a soft cue of how likely it belongs to this object. 

See Fig.~\ref{fig:matching} for an example visualization of the global matching distance map. %
It can be seen that the duck is relatively well captured, but the distance map is noisy and contains many false-positive small distances in the water. This is a strong motivation for not using these distances directly to produce segmentations but rather as an input to a segmentation head which can recover from noisy distances.

In practice we compute the global matching distance maps by a large matrix product, from which we derive all pairwise distances between the current and the first frame and then apply the object-wise minimization.

\PAR{Local Previous Frame Matching.}
In addition to transferring semantic information from the first frame using the learned embedding, we also use it to transfer information between adjacent frames to effectively enable tracking and dealing with appearance changes. Similar to the global matching distance map, we define the distance map $\hat{G}_{t,o}(p)$ with respect to the previous frame by
\begin{equation}
  \hat{G}_{t,o}(p)=\begin{cases}
  \min_{q\in\mathcal{P}_{t-1,o}}d(p,q) & \text{if }\mathcal{P}_{t-1,o}\neq\emptyset\\
  1 & \text{otherwise}
\end{cases}
\end{equation}
where the time index of $\mathcal{P}$ changed from $1$ to $t-1$. Additionally, $\mathcal{P}_{t-1,o}$ is now given by our own predictions instead of by the first-frame ground truth, which means that it can be empty, in which case we define the distance as $1$.

\begin{figure*}[t]
\centering
\includegraphics[width=0.75\textwidth]{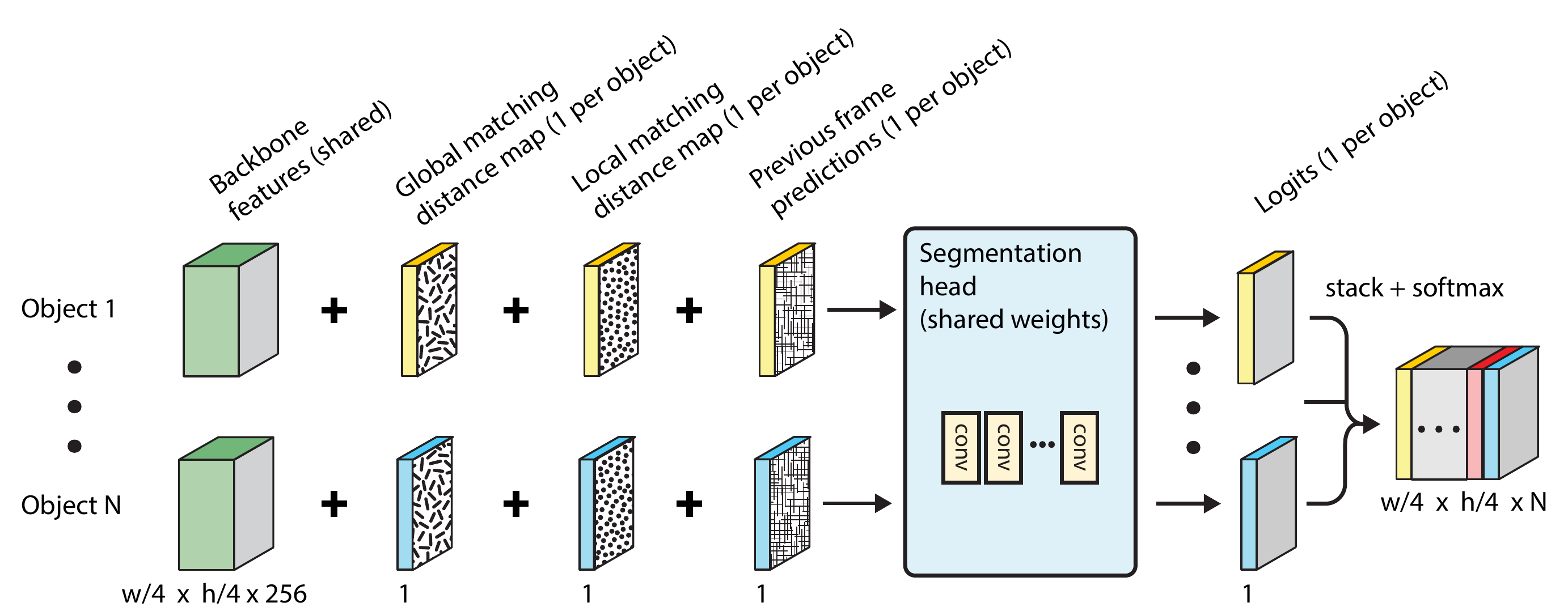}
\caption{\label{fig:seg_head} Dynamic segmentation head for systematic handling of multiple objects. The lightweight segmentation head is dynamically instantiated once for each object in the video and produces a one-dimensional feature map of logits for each object. The logits for each object are then stacked together and softmax is applied. The dynamic segmentation head can be trained with a standard cross entropy loss.}
\end{figure*}

When matching the current frame to the first frame, each pixel of the current frame needs to be compared to each pixel of the first frame, since the objects might have moved a lot over time. However, when matching with respect to the previous frame, we can exploit the fact that the motion between two frames is usually small to avoid false positive matches and save computation time. Hence, in practice we do not use $\hat{G}_{t,o}(p)$ but instead we use a local matching distance map.
Inspired by FlowNet \cite{Dosovitskiy15ICCV}, for pixel $p$ of frame $t$ we only consider pixels $q$ of frame $t-1$ %
in a local neighborhood of $p$ when searching for a nearest neighbor. For a given window size $k$, we define the neighborhood $N(p)$ as the set of pixels (regardless of which frame they are coming from) which are at most $k$ pixels away from $p$ in both $x$ and $y$ direction. This means that $N(p)$ usually comprises $(2 \cdot k + 1)^2$ elements in the same frame where $p$ is coming from, and fewer elements close to the image boundaries.
The local matching distance map $L_{t,o}(p)$ for time $t$, object $o$, and pixel $p$ is defined by
\begin{equation}
L_{t,o}(p)=\begin{cases}
\min_{q\in\mathcal{P}_{t-1,o}^{p}}d(p,q) & \text{if }\mathcal{P}_{t-1,o}^{p}\neq\emptyset\\
1 & \text{otherwise},
\end{cases}
\end{equation}
where $\mathcal{P}_{t-1,o}^{p}:=\mathcal{P}_{t-1,o}\cap N(p)$ is the set of pixels of frame $t-1$ which belong to object $o$ and are in the neighborhood of $p$.
Note that $L_{t,o}(p)$ can be efficiently computed using cross-correlation. We found that in practice using local previous frame matching, \ie,  $L_{t,o}(p)$, produces better results and is more efficient than using the global previous frame distance map $\hat{G}_{t,o}(p)$. 

Fig.~\ref{fig:matching} also shows an example visualization of a local matching distance map. Note that all pixels which are too far away from the previous frame mask are assigned a distance of 1. Since the motion between the previous and current frame was small, local matching produces a very sharp and accurate distance map.

\PAR{Previous Frame Predictions.} In addition to using our own predictions of the previous time frame for local previous frame matching, we found it helpful to use these predictions, \ie, the posterior probability map over objects, also directly as features as an additional cue.

\PAR{Dynamic Segmentation Head.}
In order to systematically and efficiently deal with a variable number of objects, we propose a dynamic  segmentation head which is dynamically instantiated once for each object with shared weights (see Fig.~\ref{fig:seg_head}). The inputs to the dynamic segmentation head for object $o$ at time $t$ are i) the global matching distance map $G_{t,o}(\cdot)$, ii) the local matching distance map $L_{t,o}(\cdot)$, iii) the probability distribution for object $o$ predicted at time $t-1$, and iv) the shared backbone features. The dynamic segmentation head consists of a few convolutional layers (see the implementation details below) which extract a one-dimensional feature map of logits for one object. Note that only three of the $259$ dimensions of the input of the dynamic segmentation head differ between distinct objects, but we found that these three dimensions in practice deliver a strong enough cue to produce accurate logits when combined with the backbone features. After for each object a one-dimensional feature map of logits is extracted, we stack them together, apply softmax over the object dimension and apply a cross entropy loss.

The segmentation head needs to be run once for each object, but the majority of the computation  occurs in extracting the shared backbone features which allows FEELVOS to scale well to multiple objects. Additionally, we are able to train end-to-end for multi-object segmentation even for a variable number of objects. Both properties are in strong contrast to many recent methods like RGMP \cite{Oh18CVPR} which evaluate a full network, designed for single-object segmentation, once for each object and heuristically combine the individual results.

\PAR{Training procedure.} Our training procedure is deliberately simple. For each training step, we first randomly select a mini-batch of videos. For each video we randomly select three frames: one frame which serves as the reference frame, \ie, it plays the role of the first frame of a video, and two adjacent frames from which the first serves as the previous frame and the second one serves as the current frame to be segmented. We apply the loss only to the current frame. During training, we use the ground truth of the previous frame for local matching and also use it to define the previous frame predictions by setting them to $1$ for the correct object and to $0$ for all other objects for each pixel. Note that our training procedure is much simpler than the one used in RGMP \cite{Oh18CVPR} which requires synthetic generation of data, and backpropagation through time.

\PAR{Inference.} Inference for FEELVOS is straightforward and requires only a single forward pass per frame. Given a test video with the ground truth for the first frame, we first extract the embedding vectors for the first frame. Afterwards, we go through the video frame-by-frame, compute the embedding vectors for the current frame, apply global matching to the first frame and local matching to the previous frame, run the dynamic segmentation head for each object, and apply a pixelwise argmax to produce the final segmentation. For the previous-frame prediction features we use the soft probability map predicted at the previous frame.

\PAR{Implementation Details.}
As the backbone of our network, we use the recent DeepLabv3+ architecture \cite{Chen18ECCV} which is based on the Xception-65 \cite{CholletCVPR17,dai2017coco} architecture and applies depthwise separable convolutions \cite{howard2017mobilenets}, batch normalization \cite{ioffe2015batch}, Atrous Spatial Pyramid Pooling \cite{chen2017deeplab, chen2017rethinking}, and a decoder module which produces features with a stride of $4$. 

On top of this we add an embedding layer consisting of one depthwise separable convolution, \ie, a $3\times3$ convolution performed separately for each channel followed by a $1\times1$ convolution, allowing interactions between channels. We extract embedding vectors of dimension $100$.

For the dynamic segmentation head, we found that a large receptive field is important. In practice, we use 4 depthwise separable convolutional layers with a dimensionality of $256$, a kernel size of $7\times7$ for the depthwise convolutions, and a ReLU activation function. On top of this we add a $1\times1$ convolution to extract logits of dimension $1$.

Computing distances for all pairs of pixels for global matching during training is expensive. We found that in practice it is unnecessary to consider all pairs. Hence, during training we randomly subsample the set of reference pixels from the first frame to contain at most $1024$ pixels per object, and found that it does not affect the results much.

For local matching we use a window size of $k=15$ applied on the embedding vectors which are extracted with a stride of $4$. We start training using weights for DeepLabv3+ which were pre-trained on ImageNet \cite{imagenet} and COCO \cite{coco}. As training data we use the DAVIS 2017 \cite{DAVIS2017} training set (60 videos) and the YouTube-VOS~\cite{Xu18ECCV} training set (3471 videos). We apply a bootstrapped cross entropy loss~\cite{bootstrappedCE,pohlen2016full} which only takes into account the $15\%$ hardest pixels for calculating the loss. We optimize using gradient descent with a momentum of $0.9$, and a learning rate of $0.0007$ for $200,\!000$ steps with a batch size of $3$ videos (\ie 9 images) per GPU using $16$ Tesla P100 GPUs. We apply  flipping and scaling as data augmentations, and crop the input images randomly to a size of $465\times465$ pixels.

\section{Experiments}
After training, our network is evaluated on the DAVIS 2016 \cite{DAVIS2016} validation set, the DAVIS 2017 \cite{DAVIS2017} validation and test-dev sets, and the YouTube-Objects \cite{YoutubeObjectsOriginal,YoutubeObjectsSegmentation} dataset.
The DAVIS 2016 validation set consists of 20 videos for each of which a single instance is annotated.
The DAVIS 2017 dataset comprises a training set of 60 sequences with multiple annotated instances and a validation set that extends the DAVIS 2016 validation set to a total of 30 videos with multiple instances annotated. The DAVIS 2017 test-dev set also contains 30 sequences. The YouTube-Objects dataset \cite{YoutubeObjectsOriginal,YoutubeObjectsSegmentation} consists of 126 videos with sparse annotations of a single instance per video. 

We adopt the evaluation measures defined by DAVIS \cite{DAVIS2016}. The first evaluation criterion is the mean intersection-over-union (mIoU) between the predicted and the ground truth segmentation masks, denoted by $\mathcal{J}$. The second evaluation criterion is the contour accuracy $\mathcal{F}$, described in more detail in \cite{DAVIS2016}. Finally, $\mathcal{J}$\&$\mathcal{F}$ is the average of $\mathcal{J}$ and $\mathcal{F}$.

{\footnotesize{}}
\begin{table}[t]
\centering{}{\footnotesize{}}%
\scalebox{0.92}{
\begin{tabular}{cccccc}
\toprule 
 & {\footnotesize{}FT} & {\footnotesize{}$\mathcal{J}$} & {\footnotesize{}$\mathcal{F}$} & {\footnotesize{}$\mathcal{J}$\&$\mathcal{F}$} & {\footnotesize{}t/s}\tabularnewline
\midrule
{\footnotesize{}OSMN \cite{Yang18CVPR}} &  & {\footnotesize{}$52.5$} & {\footnotesize{}$57.1$} & {\footnotesize{}$54.8$} & \textbf{\footnotesize{}$\mathbf{0.28^{\dagger}}$}\tabularnewline
{\footnotesize{}FAVOS \cite{Cheng18CVPR}} &  & {\footnotesize{}$54.6$} & {\footnotesize{}$61.8$} & {\footnotesize{}$58.2$} & {\footnotesize{}$1.2^{\dagger}$}\tabularnewline
{\footnotesize{}VideoMatch \cite{Hu18ECCV}} &  & {\footnotesize{}$56.5$} & {\footnotesize{}$68.2$} & {\footnotesize{}$62.4$} & {\footnotesize{}$0.35$}\tabularnewline
{\footnotesize{}RGMP \cite{Oh18CVPR}} &  & {\footnotesize{}$64.8$} & {\footnotesize{}$68.6$} & {\footnotesize{}$66.7$} & \textbf{\footnotesize{}$\mathbf{0.28^{\dagger}}$}\tabularnewline
{\footnotesize{}FEELVOS (ours, -YTB-VOS)} &  & \textbf{\footnotesize{}$65.9$} & \textbf{\footnotesize{}$72.3$} & \textbf{\footnotesize{}$69.1$} & {\footnotesize{}$0.51$}\tabularnewline
{\footnotesize{}FEELVOS (ours)} &  & {\footnotesize{}$\mathbf{69.1}$} & {\footnotesize{}$\mathbf{74.0}$} & {\footnotesize{}$\mathbf{71.5}$} & {\footnotesize{}$0.51$}\tabularnewline
\midrule 
{\footnotesize{}OnAVOS \cite{voigtlaender17BMVC}} & {\footnotesize{}\ding{51}} & {\footnotesize{}$61.0$} & {\footnotesize{}$66.1$} & {\footnotesize{}$63.6$} & {\footnotesize{}$\mathbf{26}$}\tabularnewline
{\footnotesize{}PReMVOS \cite{Luiten18ACCV}} & {\footnotesize{}\ding{51}} & {\footnotesize{}$\mathbf{73.9}$} & {\footnotesize{}$\mathbf{81.7}$} & {\footnotesize{}$\mathbf{77.8}$} & {\footnotesize{}$37.6$}\tabularnewline
\bottomrule
\end{tabular}}{\footnotesize{}\caption{\label{tab:results-davis17}Quantitative results on the DAVIS 2017
validation set. FT denotes fine-tuning, and t/s denotes time per frame
in seconds. $\dagger$: timing extrapolated from DAVIS 2016 assuming
linear scaling in the number of objects.}
}{\footnotesize \par}
\end{table}
{\footnotesize \par}

{\footnotesize{}}
\begin{table}[t]
\centering{}{\footnotesize{}}%
\scalebox{0.92}{
\begin{tabular}{cccccc}
\toprule 
 & {\footnotesize{}FT} & {\footnotesize{}$\mathcal{J}$} & {\footnotesize{}$\mathcal{F}$} & {\footnotesize{}$\mathcal{J}$\&$\mathcal{F}$} & {\footnotesize{}t/s}\tabularnewline
\midrule
{\footnotesize{}RGMP \cite{Oh18CVPR}} &  & {\footnotesize{}$51.4$} & {\footnotesize{}$54.4$} & {\footnotesize{}$52.9$} & {\footnotesize{}$\mathbf{0.42^{\dagger}}$}\tabularnewline
{\footnotesize{}FEELVOS (ours, -YTB-VOS)} &  & {\footnotesize{}$51.2$} & {\footnotesize{}$57.5$} & {\footnotesize{}$54.4$} & {\footnotesize{}$0.54$}\tabularnewline
{\footnotesize{}FEELVOS (ours)} &  & {\footnotesize{}$\mathbf{55.2}$} & {\footnotesize{}$\mathbf{60.5}$} & {\footnotesize{}$\mathbf{57.8}$} & {\footnotesize{}$0.54$}\tabularnewline
\midrule 
{\footnotesize{}OnAVOS \cite{voigtlaender17BMVC,voigtlaender17DAVIS}} & {\footnotesize{}\ding{51}} & {\footnotesize{}$53.4$} & {\footnotesize{}$59.6$} & {\footnotesize{}$56.5$} & {\footnotesize{}$\mathbf{39}$}\tabularnewline
{\footnotesize{}PReMVOS \cite{Luiten18ACCV}} & {\footnotesize{}\ding{51}} & {\footnotesize{}$\mathbf{67.5}$} & {\footnotesize{}$\mathbf{75.7}$} & {\footnotesize{}$\mathbf{71.6}$} & {\footnotesize{}$41.3$}\tabularnewline
\bottomrule
\end{tabular}}{\footnotesize{}\caption{\label{tab:results-davis17-testdev}Quantitative results on the DAVIS
2017 test-dev set. FT denotes fine-tuning, and t/s denotes time per
frame in seconds. $\dagger$: timing extrapolated from DAVIS 2016
assuming linear scaling in the number of objects.}
}
\end{table}
{\footnotesize \par}

\begin{figure}[t]
\centering
\resizebox{\linewidth}{!}{\begin{tikzpicture}[/pgfplots/width=1\linewidth, /pgfplots/height=0.64\linewidth, /pgfplots/legend pos=south east]
    \begin{axis}[ymin=0.53,ymax=0.80,xmin=0.15,xmax=50,enlargelimits=false,
        xlabel=Time per frame (seconds),
        ylabel=Region and contour quality ($\mathcal{J}$\&$\mathcal{F}$),
		font=\scriptsize,
        grid=both,
		grid style=dotted,
        xlabel shift={-2pt},
        ylabel shift={-5pt},
        xmode=log,
        legend columns=1,
        minor ytick={0,0.025,...,1.1},
        ytick={0,0.1,...,1.1},
	    yticklabels={0,.1,.2,.3,.4,.5,.6,.7,.8,.9,1},
	    xticklabels={.1,1,10,100},
        legend pos= outer north east,
        legend cell align={left}
        ]
        
        \addplot[red,mark=*,only marks,line width=0.75, mark size=2.3] coordinates{(0.51,0.715)};
        \addlegendentry{\hphantom{i}FEELVOS (ours)}
        
   	    \addplot[green,mark=*,only marks,line width=0.75] coordinates{(0.51,0.691)};
        \addlegendentry{\hphantom{i}FEELVOS {\tiny{}(-YTB-VOS)}}
        
        \addplot[blue,mark=+,only marks,line width=0.75] coordinates{(37.4,0.778)};
        \addlegendentry{\hphantom{i}PReMVOS \cite{Luiten18ACCV, Luiten18DAVIS}}

        \addplot[magenta,mark=square,only marks,line width=0.75, mark size=1.45] coordinates{(0.35,0.624)};
        \addlegendentry{\hphantom{i}VideoMatch \cite{Hu18ECCV}}

        \addplot[black,mark=square,only marks,line width=0.75, mark size=1.45] coordinates{(26,	0.636)};
        \addlegendentry{\hphantom{i}OnAVOS \cite{voigtlaender17BMVC, voigtlaender17DAVIS}}
        
        \addplot[olive,mark=asterisk, mark size=1.9,only marks, line width=0.75] coordinates{(9,0.68)};
        \addlegendentry{\hphantom{i}OSVOS-S \cite{Maninis18TPAMI}}
        
        \addplot[magenta,mark=x, mark size=2.1,only marks, line width=0.75]coordinates{(4.66,0.741)};
        \addlegendentry{\hphantom{i}DyeNet \cite{Li18ECCV}}
                
        \addplot[cyan,mark=star, mark size=2,only marks, line width=0.75] coordinates{(0.26,0.667)};
        \addlegendentry{\hphantom{i}RGMP \cite{Oh18CVPR}}
                
        \addplot[green,mark=oplus, mark size=1.6,only marks, line width=0.75] coordinates{(0.28,0.548)};
        \addlegendentry{\hphantom{i}OSMN \cite{Yang18CVPR}}
                
        \addplot[black,mark=Mercedes star, mark size=2.2,only marks, line width=0.75] coordinates{(3.6,0.582)};
        \addlegendentry{\hphantom{i}FAVOS \cite{Cheng18CVPR}}
    \end{axis}
\end{tikzpicture}}
\vspace{-5mm}
   \caption{Quality versus timing on DAVIS 2017. The proposed FEELVOS shows a very good speed/accuracy trade-off. FEELVOS (-YTB-VOS) denotes training without YouTube-VOS \cite{Xu18ECCV}  training data.}
   \label{fig:qual_vs_time}
   \vspace{-3mm}
\end{figure}
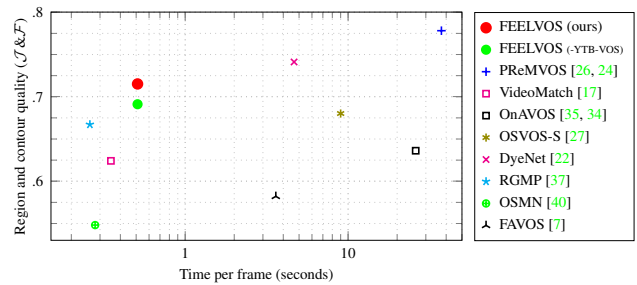

{\footnotesize{}}
\begin{table}[t]
\centering{}{\footnotesize{}}%
\scalebox{0.92}{
\begin{tabular}{cccccc}
\toprule 
 & {\footnotesize{}FT} & {\footnotesize{}$\mathcal{J}$} & {\footnotesize{}$\mathcal{F}$} & {\footnotesize{}$\mathcal{J}$\&$\mathcal{F}$} & {\footnotesize{}t/s}\tabularnewline
\midrule
{\footnotesize{}OSMN \cite{Yang18CVPR}} &  & {\footnotesize{}$74.0$} & {\footnotesize{}-} & {\footnotesize{}-} & {\footnotesize{}$\mathbf{0.14}$}\tabularnewline
{\footnotesize{}FAVOS \cite{Cheng18CVPR}} &  & {\footnotesize{}$77.9$} & {\footnotesize{}$76.0$} & {\footnotesize{}$77.0$} & {\footnotesize{}$0.6$}\tabularnewline
{\footnotesize{}PML \cite{Chen18CVPR}} &  & {\footnotesize{}$75.5$} & {\footnotesize{}$79.3$} & {\footnotesize{}$77.4$} & {\footnotesize{}$0.28$}\tabularnewline
{\footnotesize{}VideoMatch \cite{Hu18ECCV}} &  & {\footnotesize{}$81.0$} & {\footnotesize{}$80.8$} & {\footnotesize{}$80.9$} & {\footnotesize{}$0.32$}\tabularnewline
{\footnotesize{}RGMP \cite{Oh18CVPR} (-sim. data)} &  & {\footnotesize{}$68.6$} & {\footnotesize{}$68.9$} & {\footnotesize{}$68.8$} & \textbf{\footnotesize{}$\mathbf{0.14}$}\tabularnewline
{\footnotesize{}RGMP \cite{Oh18CVPR}} &  & \textbf{\footnotesize{}$\mathbf{81.5}$} & {\footnotesize{}$82.0$} & \textbf{\footnotesize{}$\mathbf{81.8}$} & \textbf{\footnotesize{}$\mathbf{0.14}$}\tabularnewline
{\footnotesize{}FEELVOS (ours, -YTB-VOS)} &  & {\footnotesize{}$80.3$} & \textbf{\footnotesize{}$\mathbf{83.1}$} & \textbf{\footnotesize{}$\mathbf{81.7}$} & {\footnotesize{}$0.45$}\tabularnewline
{\footnotesize{}FEELVOS (ours)} &  & {\footnotesize{}$81.1$} & {\footnotesize{}$82.2$} & \textbf{\footnotesize{}$\mathbf{81.7}$} & {\footnotesize{}$0.45$}\tabularnewline
\midrule 
{\footnotesize{}OnAVOS \cite{voigtlaender17BMVC}} & {\footnotesize{}\ding{51}} & {\footnotesize{}$\mathbf{85.7}$} & {\footnotesize{}$84.2$} & {\footnotesize{}$85.0$} & {\footnotesize{}$\mathbf{13}$}\tabularnewline
{\footnotesize{}PReMVOS \cite{Luiten18ACCV}} & {\footnotesize{}\ding{51}} & {\footnotesize{}$84.9$} & {\footnotesize{}$\mathbf{88.6}$} & {\footnotesize{}$\mathbf{86.8}$} & {\footnotesize{}$32.8$}\tabularnewline
\bottomrule
\end{tabular}}{\footnotesize{}\caption{\label{tab:results-davis16}Quantitative results on the DAVIS 2016
validation set. FT denotes fine-tuning, and t/s denotes time per frame
in seconds.}
}
\end{table}
{\footnotesize \par}

{\footnotesize{}}
\begin{table}[t]
\centering{}{\footnotesize{}}%
\begin{tabular}{ccc}
\toprule 
 & {\footnotesize{}FT} & {\footnotesize{}$\mathcal{J}$}\tabularnewline
\midrule
{\footnotesize{}OSMN \cite{Yang18CVPR}} &  & {\footnotesize{}$69.0^{*}$}\tabularnewline
{\footnotesize{}VideoMatch \cite{Hu18ECCV}} &  & {\footnotesize{}$79.7^{*}$}\tabularnewline
{\footnotesize{}FEELVOS (ours, -YTB-VOS)} &  & {\footnotesize{}$78.9$}\tabularnewline
{\footnotesize{}FEELVOS (ours)} &  & {\footnotesize{}$\mathbf{82.1}$}\tabularnewline
\midrule 
{\footnotesize{}MaskTrack \cite{masktrack}} & {\footnotesize{}\ding{51}} & {\footnotesize{}$77.7$}\tabularnewline
{\footnotesize{}OSVOS \cite{OSVOS}} & {\footnotesize{}\ding{51}} & {\footnotesize{}$78.3$}\tabularnewline
{\footnotesize{}OnAVOS \cite{voigtlaender17BMVC}} & {\footnotesize{}\ding{51}} & {\footnotesize{}$\mathbf{80.5}$}\tabularnewline
\bottomrule
\end{tabular}{\footnotesize{}\caption{\label{tab:results-youtube-objects}Quantitative results on the YouTube-Objects
dataset. FT denotes fine-tuning. {*}: The used evaluation protocol
for YouTube-Objects is inconsistent, \eg sometimes the first frame is ignored; the results
marked with {*} might not be directly comparable.%
}
}
\end{table}
{\footnotesize \par}

\PAR{Main Results.} Tables~\ref{tab:results-davis17} and \ref{tab:results-davis17-testdev} compare our results on the DAVIS 2017 validation and test-dev sets to recent other methods which do not employ fine-tuning and to PReMVOS \cite{Luiten18ACCV, Luiten18DAVIS, Luiten18ECCVW} and OnAVOS \cite{voigtlaender17BMVC}. Note that PML \cite{Chen18CVPR} does not provide results for DAVIS 2017. 
On the validation set, FEELVOS improves over the previously best non-finetuning method RGMP \cite{Oh18CVPR} both in terms of mIoU $\mathcal{J}$ and contour accuracy $\mathcal{F}$. For non-fine-tuning methods FEELVOS achieves a new state of the art with a $\mathcal{J}$\&$\mathcal{F}$ score of $71.5\%$ which is $4.8\%$ higher than RGMP and $2.4\%$ higher when not using YouTube-VOS data for training (denoted by -YTB-VOS). %
FEELVOS's result is stronger than the result of OnAVOS \cite{voigtlaender17BMVC} which heavily relies on fine-tuning. However, FEELVOS cannot match the results of the heavily engineered and slow PReMVOS \cite{Luiten18ACCV}. %
On the DAVIS 2017 test-dev set, FEELVOS achieves a $\mathcal{J}$\&$\mathcal{F}$ score of $57.8\%$, which is $4.9\%$ higher than the result of RGMP \cite{Oh18CVPR}. Here, the runtime of RGMP and FEELVOS is almost identical since FEELVOS's runtime is almost independent of the number of objects and the test-dev set contains more objects per sequence.%

Fig.~\ref{fig:qual_vs_time} shows the $\mathcal{J}$\&$\mathcal{F}$ score and runtime of current methods with and without fine-tuning. It can be seen that FEELVOS achieves a very good speed/accuracy trade-off with a runtime of $0.51$ seconds per frame.

Table~\ref{tab:results-davis16} shows the results on the simpler DAVIS 2016 validation set which only has a single object instance annotated per sequence. Here, FEELVOS's result with a $\mathcal{J}$\&$\mathcal{F}$ score of $81.7\%$ is comparable to RGMP, which achieves $81.8\%$. However, RGMP heavily relies on simulated training data, which our method does not require. Without the simulated data, RGMP achieves only $68.8\%$ $\mathcal{J}$\&$\mathcal{F}$. Again, FEELVOS is not able to reach the results of fine-tuning based methods, but it achieves a very good speed/accuracy trade-off and only uses a single neural network.

Table~\ref{tab:results-youtube-objects} shows the results on YouTube-Objects \cite{YoutubeObjectsOriginal,YoutubeObjectsSegmentation}. Note that the evaluation protocol for this dataset is not always consistent and the results marked with * might not be directly comparable. FEELVOS achieves a $\mathcal{J}$ score of $82.1\%$ which is even better than the results of the fine-tuning based methods OSVOS \cite{OSVOS} and OnAVOS \cite{voigtlaender17BMVC}.

\begin{table}
\begin{centering}
{\footnotesize{}}%
\begin{tabular}{cccccccc}
\toprule 
 & {\footnotesize{}FF-GM} & {\footnotesize{}PF-LM} & {\footnotesize{}PF-GM} & {\footnotesize{}PFP} & {\footnotesize{}$\mathcal{J}$} & {\footnotesize{}$\mathcal{F}$} & {\footnotesize{}$\mathcal{J}\&\mathcal{F}$}\tabularnewline
\midrule
{\footnotesize{}1}%
{} & {\footnotesize{}\ding{51}} & {\footnotesize{}\ding{51}} &  & {\footnotesize{}\ding{51}} & {\footnotesize{}$65.9$} & {\footnotesize{}$72.3$} & {\footnotesize{}$69.1$}\tabularnewline
\midrule 
{\footnotesize{}2}%
{} & {\footnotesize{}\ding{51}} &  & {\footnotesize{}\ding{51}} & {\footnotesize{}\ding{51}} & {\footnotesize{}$61.2$} & {\footnotesize{}$67.3$} & {\footnotesize{}$64.2$}\tabularnewline
{\footnotesize{}3}%
{} & {\footnotesize{}\ding{51}} &  &  & {\footnotesize{}\ding{51}} & {\footnotesize{}$49.9$} & {\footnotesize{}$59.8$} & {\footnotesize{}$54.9$}\tabularnewline
{\footnotesize{}4}%
{} & {\footnotesize{}\ding{51}} &  &  &  & {\footnotesize{}$47.3$} & {\footnotesize{}$57.9$} & {\footnotesize{}$52.6$}\tabularnewline
{\footnotesize{}5}%
{} & {\footnotesize{}\ding{51}} & {\footnotesize{}\ding{51}} &  &  & {\footnotesize{}$60.4$} & {\footnotesize{}$66.2$} & {\footnotesize{}$63.3$}\tabularnewline
{\footnotesize{}6}%
{} &  & {\footnotesize{}\ding{51}} &  & {\footnotesize{}\ding{51}} & {\footnotesize{}$53.8$} & {\footnotesize{}$58.3$} & {\footnotesize{}$56.1$}\tabularnewline
\bottomrule
\end{tabular}
\par\end{centering}{\footnotesize \par}
\caption{\label{tab:ablation-study}Ablation study on DAVIS 2017. FF and PF denote first
frame and previous frame, respectively, and GM and LM denote global
matching and local matching. PFP denotes using the previous frame
predictions as input to the dynamic segmentation head.%
}
\end{table}

\PAR{Ablation Study.} In Table~\ref{tab:ablation-study} we analyze the effect of the individual components of FEELVOS on the DAVIS 2017 validation set. For simplicity, we only use the smaller DAVIS 2017 training set as training data for these experiments.
Line 1 is the proposed FEELVOS which uses first-frame global matching (FF-GM), previous frame local matching (PV-LM), and previous frame predictions (PFP) as inputs for the dynamic segmentation head. This setup achieves a $\mathcal{J}$\&$\mathcal{F}$ score of $69.1\%$.

\begin{figure*}[t]
\includegraphics[width=1.0\textwidth]{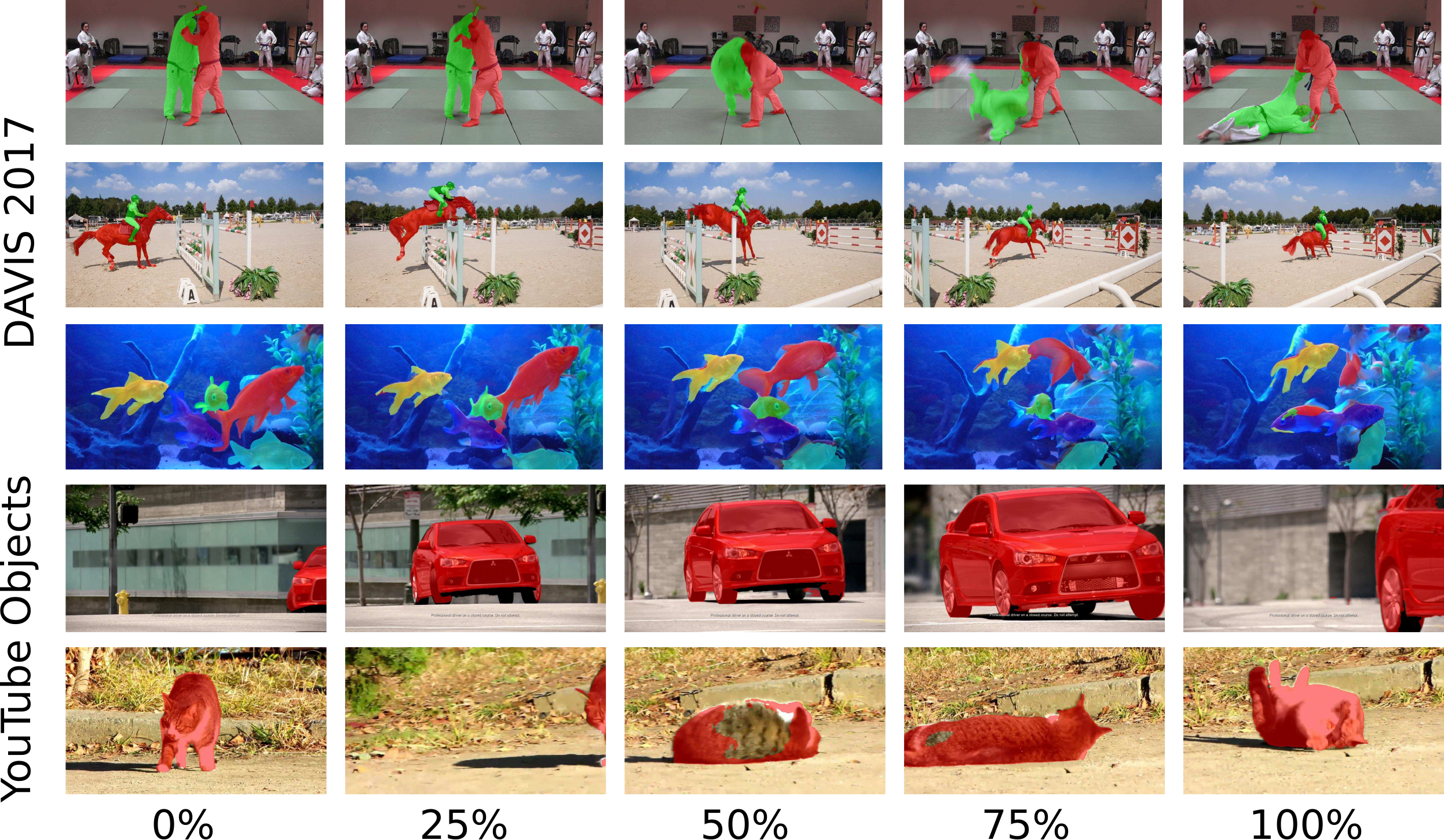}
\caption{\label{fig:qualitative}Qualitative results on the DAVIS 2017 validation set and on YouTube-Objects. In the third row the many similar looking fishes cause FEELVOS to lose track of some of them. In the last row, FEELVOS fails to segment some of the back of the cat.
}
\end{figure*}

In line 2, we replace the previous frame local matching (PF-LM) by previous frame global matching (PF-LM), which significantly degrades the results by almost $5\%$ to $64.2\%$ and shows the effectiveness of restricting the previous frame matching to a local window.

In line 3, we disable previous frame matching completely. Here the results drop even more to $54.9\%$ which shows that matching to the previous frame using the learned embedding is extremely important to achieve good results.

In line 4, we additionally disable the use of previous frame predictions (PFP) as features to the dynamic segmentation head. In this setup, each frame can be segmented individually and only information from matching globally to the first frame is used. In this case, the results deteriorate even further to $52.6\%$.

In line 5, we use previous frame local matching (PF-LM) again, but disable the use of previous frame predictions (PFP). In this case, the result is $63.3\%$ which is much better than the result of line 3 which used PFP but no PF-LM. This shows that previous frame local matching is a more effective way to transfer information from the previous frame than just using the previous frame predictions as features. It also shows that both ways to transfer information over time are complementary and their combination is most effective.

In line 6, we use PF-LM and PFP but disable the first-frame global matching. This means that the first-frame information is only used to initialize the mask used for PF-LM and PFP but no longer as explicit guidance for each frame. Here the result deteriorates compared to line 1 by $13\%$ which shows that matching to the first frame is extremely important to achieve good results.

In summary, %
we showed that each component of FEELVOS is useful, that matching in embedding space to the previous frame is extremely effective, and that the proposed local previous frame matching performs significantly better than globally matching to the previous frame.

\PAR{Qualitative Results.}
Fig.~\ref{fig:qualitative} shows qualitative results of FEELVOS on the DAVIS 2017 validation set and the YouTube-Objects dataset. It can be seen that in many cases FEELVOS is able to produce accurate segmentations even in difficult cases like large motion in the judo sequence or a truncated first frame for the car. In the challenging fish sequence (third row), FEELVOS looses track of some of the fishes, probably because their appearance is very similar. In the last row, FEELVOS fails to segment some parts of the back of the cat. This is most likely because the back texture was not seen in the first frame. However, afterwards, FEELVOS is able to recover from that error.%
\vspace{-1pt}

\section{Conclusion}

We started with the observation that there are many strong methods for VOS, but many of them lack practical usability. Based on this insight, we defined several design goals which a practical useful method for VOS should fulfill. Most importantly we aim for a fast and simple method which yet achieves strong results. To this end, we propose FEELVOS which learns a semantic embedding for segmenting multiple objects in an end-to-end way. The key components of FEELVOS are global matching to the first frame of the video and local matching to the previous frame. We showed experimentally that each component of FEELVOS is highly effective and we achieve new state of the art results on DAVIS 2017 for VOS without fine-tuning. Overall, FEELVOS is a fast and practical useful method for VOS and we hope that our work will inspire more methods which fulfill the design criteria defined by us and advance the state of the art for practical useful methods in VOS.

\footnotesize \PAR{Acknowledgements:} We would like to thank Alireza Fathi, Andre Araujo, Bryan Seybold, Jonathon Luiten, and Jordi Pont-Tuset for helpful discussions and Yukun Zhu for help with open-sourcing our models.

{\small
\bibliographystyle{ieee}
\bibliography{abbrev_short,paper}
}

\end{document}